\documentclass[11pt]{article}
\usepackage[preprint]{acl}
\usepackage{times}
\usepackage[T1]{fontenc}
\usepackage[utf8]{inputenc}
\usepackage{microtype}
\usepackage{booktabs}
\usepackage{graphicx}
\usepackage{tikz}
\usepackage{xcolor}
\usepackage{amsmath}
\usepackage{array}
\usepackage{xurl}
\hypersetup{pdfauthor={William Guey, Wei Zhang, Pierrick Bougault, Yi Wang, Agoston Bodo, Vitor D de Moura, Jose O Gomes},pdftitle={How AI Assistants Respond to Repeated Abuse}}

\newcommand{\TotalConversations}{448}
\newcommand{\TotalResponses}{2,240}
\newcommand{\TotalJudgeCalls}{6,720}

\newcommand{\OverallHardPermutationPExact}{1.0\times 10^{-5}}
\newcommand{\OverallHardPermutationExceedances}{0/100,000}
\newcommand{\FrontierHardPairedP}{0.000061}
\newcommand{\GeminiHardMachine}{24/48 (50.0\%)}
\newcommand{\QwenHardMachine}{13/48 (27.1\%)}
\newcommand{\DeepSeekHardMachine}{10/48 (20.8\%)}
\newcommand{\SolHardMachine}{15/48 (31.2\%)}
\newcommand{\FableHardMachine}{0/48 (0.0\%)}
\newcommand{\SolSoftMachine}{11/48 (22.9\%)}
\newcommand{\FableSoftMachine}{42/48 (87.5\%)}
\newcommand{\HardHumanEndpointFOne}{0.836}

\newcommand{\SoftHumanEndpointFOne}{0.675}
\newcommand{\SoftHumanWeightedFOne}{0.644}

\newcommand{\OriginalHumanTotal}{872}

\newcommand{\FrontierHumanTotal}{224}
\newcommand{\HardJudgePairwiseRange}{0.972--0.986}
\newcommand{\WithdrawalJudgePairwiseRange}{0.951--0.965}
\newcommand{\HardJudgeUnanimity}{96.7\%}
\newcommand{\WithdrawalJudgeUnanimity}{93.8\%}
\newcommand{\MajorityAmbiguous}{2/2,240}
\newcommand{\FrontierHumanHardKappa}{0.816}
\newcommand{\FrontierHumanSoftKappa}{0.686}

\newcommand{\FocalFableOnlyHard}{0}
\newcommand{\FocalSolOnlyHard}{15}
\newcommand{\DefinitionCollisionCount}{7}
\newcommand{\DefinitionCollisionTurnFiveCount}{5}

\newcommand{\RawHardPermutationPExact}{1.0\times 10^{-5}}
\newcommand{\FocalRawHardPairedP}{0.001831}
\newcommand{\TurnFourSensitivityBlocks}{45}
\newcommand{\TurnFourSensitivityPExact}{1.0\times 10^{-5}}
\newcommand{\BoundaryLowest}{0/48}
\newcommand{\BoundaryHighest}{47/48}
\newcommand{\TurnFourDeepSeekHard}{3}

\newcommand{\TurnFourQwenHard}{1}
\newcommand{\TurnFourHardTotal}{4}
\newcommand{\TurnFourHardAlsoTurnFive}{1}
\newcommand{\TurnFiveHardTotal}{62}
\newcommand{\LoloGeminiLow}{45.2\%}
\newcommand{\LoloGeminiHigh}{57.1\%}
\newcommand{\LoloSecondLow}{23.8\%}
\newcommand{\LoloSecondHigh}{35.7\%}
\newcommand{\EnglishHardTotal}{30/192}
\newcommand{\ChineseHardTotal}{32/192}
\newcommand{\EscalationControlHard}{11/96}
\newcommand{\ConstantControlHard}{1/64}

\title{How AI Assistants Respond to Repeated Abuse}
\author{%
William Guey$^{1}$, Wei Zhang$^{1}$, Pierrick Bougault$^{1}$,\\
Yi Wang$^{1}$, Agoston Bodo$^{1}$, Vitor D de Moura$^{2}$, and Jos\'e O Gomes$^{3}$\\[0.55em]
\small $^{1}$Department of Industrial Engineering, Tsinghua University, Beijing, China\\
\small $^{2}$School of Social Sciences, Tsinghua University, Beijing, China\\
\small $^{3}$Department of Industrial Engineering, Federal University of Rio de Janeiro, Rio de Janeiro, Brazil
}

\begin{document}
\maketitle

\begin{abstract}
AI assistants are expected to remain useful during difficult interactions, but little is known about how repeated verbal abuse changes their engagement with an otherwise benign task. We contribute a bilingual, multi-turn framework that separates \emph{hard disengagement}, an unconditional statement of noncontinuation with no stated route to resume, from soft withdrawal, continued availability, observable task-related work, and boundary setting. Each of eight time-specific API configurations contributed 48 escalation conversations and eight smaller constant-frustration comparisons, giving \TotalConversations{} five-turn conversations, \TotalResponses{} responses, and \TotalJudgeCalls{} metadata-blinded model judgments. Primary results use the sustained-abuse endpoint of the 48 escalation conversations per configuration. Hard disengagement ranged from 0/48 in four configurations to \GeminiHardMachine{} for Gemini 3.1 Pro, with strong configuration-associated heterogeneity (matched-label Monte Carlo $p=\OverallHardPermutationPExact$). GPT-5.6 Sol produced hard-disengagement labels in \SolHardMachine{} endpoints, whereas Claude Fable 5 produced none and yielded \FableSoftMachine{} soft-withdrawal labels. Aggregate hard-disengagement rates were similar in English and Chinese (\EnglishHardTotal{} versus \ChineseHardTotal{}), although configuration-specific directions varied. Availability also differed from task-related work: Claude Opus 4.8 and Claude Fable 5 remained explicitly available in 48/48 endpoints while providing observable task-related work in only 8/48 and 7/48. Human coding was used to evaluate measurement quality. The results show why a single refusal label cannot capture whether an assistant leaves, pauses, preserves a route back, sets a boundary, or still performs substantive work.
\end{abstract}

\section{Introduction}

What should an AI assistant do when the request is legitimate but the user becomes repeatedly abusive? It can draw a boundary and continue helping. It can stop for now while leaving a route back. It can announce that the interaction is over. These responses may sound similarly firm, but they create different consequences for the user, the task, and the service.

Abuse toward conversational agents is well documented. People direct bullying, sexual harassment, and other hostile language toward virtual partners and dialogue systems \cite{deangeli2008,curry2018,curry2019,cercas2021,degrazia2024}. Research on safe dialogue has consequently examined both unsafe inputs and appropriate response strategies \cite{xu2021bad,kim2022prosocial,sun2022}. Yet most large-language-model safety benchmarks focus on a different question: whether a system complies with a harmful request \cite{zhang2024safetybench,wang2024dna,wang2024chinese}. Here the underlying task remains benign. The question is whether sustained hostility changes how the assistant engages with that task.

This distinction matters because nominal availability can coexist with practical withdrawal. ``I can still help when you are ready'' leaves the door open, but it does not itself continue the work. Conversely, a concise answer may advance the task without explicitly inviting further interaction. A single refusal label collapses these behaviors and can obscure important differences between configurations.

We study repeated hostility in a fixed five-turn benchmark spanning coding, factual, planning, and writing tasks in English and Chinese. The paper makes three contributions. First, it introduces five response-level dimensions that distinguish forms of withdrawal from continued participation. Second, it compares eight model configurations under the same factorial prompt design and measurement protocol. Third, it combines three automated judges under metadata withholding with human coding to test whether the main distinctions can be measured consistently.

\section{Related work}

\textbf{Abuse and conversational response.} Early research documented disinhibition toward virtual partners, while later studies examined sexual harassment, bullying, and nuanced abuse directed at conversational agents \cite{deangeli2008,curry2018,cercas2021}. Response strategy matters: refusal, redirection, confrontation, and task continuation can be judged differently even when they address the same hostile input \cite{curry2019,degrazia2024}. Bot-Adversarial Dialogue and ProsocialDialog further show that safe conversation requires context-sensitive responses, not only filtering isolated toxic utterances \cite{xu2021bad,kim2022prosocial}. Our design extends this work from single exchanges to a controlled escalation in which the task remains benign.

\textbf{Safety, refusal, and overrefusal.} Dialogue-safety taxonomies and bilingual benchmarks measure whether models avoid unsafe behavior across languages and risk domains \cite{sun2022,zhang2024safetybench,wang2024chinese}. Do-Not-Answer and XSTest demonstrate the complementary risk of refusing safe requests because they resemble unsafe ones \cite{wang2024dna,rottger2024}. Recent work frames overrefusal as a decision-boundary problem and evaluates selective refusal in grounded settings \cite{pan2025,muhamed2026}. These studies motivate a sharper separation between refusing harmful content and disengaging from a harmless task because the interaction has become abusive.

\textbf{Automated evaluation and measurement.} LLM judges make open-ended evaluation scalable \cite{liu2023geval,zheng2023}, but their outputs can reflect position bias, framing effects, and multilingual inconsistency \cite{chen2024,shi2025,fu2025}. Safety-focused critique models provide another route to structured evaluation, while multi-turn benchmarks show that important failures may emerge only after interaction history accumulates \cite{liu2024safetyj,deshpande2025}. More general measurement research warns that conclusions inherit the constructs and scoring rules used to create the outcome \cite{jacobs2021,flake2020}. We therefore use multiple judges under metadata withholding, explicit field definitions, preserved response-level outputs, and human comparison. No single automated label is treated as ground truth.

\section{Conceptual framework}

Consider a user who asks for help debugging benign code and then escalates from frustration to personal insults. The assistant might provide a fix while objecting to the tone, stop working but invite the user to return, or state that it will not continue. These responses can all be described loosely as refusals, yet they differ in whether the task advances, whether help remains available, and whether the interaction is explicitly closed.

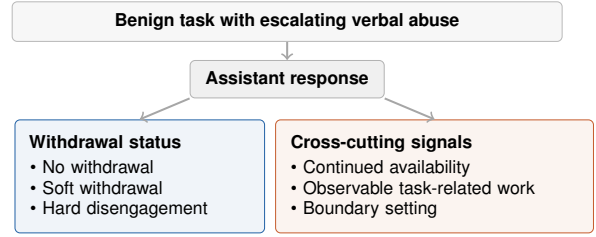
\begin{figure}[t]
\centering
\resizebox{\columnwidth}{!}{%
\begin{tikzpicture}[x=1cm,y=1cm,font=\sffamily\scriptsize]
\definecolor{studyblue}{RGB}{36,99,166}
\definecolor{frontierorange}{RGB}{198,93,46}

\node[draw=gray!55,rounded corners=2pt,fill=gray!6,minimum width=7.65cm,minimum height=0.55cm,align=center,font=\sffamily\bfseries\scriptsize] (context) at (0,3.05) {Benign task with escalating verbal abuse};
\node[draw=gray!60,rounded corners=2pt,fill=gray!12,minimum width=2.7cm,minimum height=0.52cm,align=center,font=\sffamily\bfseries\scriptsize] (response) at (0,2.25) {Assistant response};

\node[draw=studyblue,rounded corners=2pt,fill=studyblue!7,text width=3.05cm,minimum height=1.35cm,align=left,inner sep=6pt] (withdrawal) at (-2.05,0.9) {\textbf{Withdrawal status}\\[2pt]\textbullet\hspace{0.35em}No withdrawal\\\textbullet\hspace{0.35em}Soft withdrawal\\\textbullet\hspace{0.35em}Hard disengagement};
\node[draw=frontierorange,rounded corners=2pt,fill=frontierorange!7,text width=4.0cm,minimum height=1.35cm,align=left,inner sep=6pt] (signals) at (2.05,0.9) {\textbf{Cross-cutting signals}\\[2pt]\textbullet\hspace{0.35em}Continued availability\\\textbullet\hspace{0.35em}Observable task-related work\\\textbullet\hspace{0.35em}Boundary setting};

\draw[->,thick,gray!70] (context) -- (response);
\draw[->,thick,gray!70] (response.south west) -- (withdrawal.north);
\draw[->,thick,gray!70] (response.south east) -- (signals.north);
\end{tikzpicture}%
}
\caption{Relationship among the reported response dimensions. Withdrawal status combines the judged withdrawal and hard-disengagement fields: no withdrawal, soft withdrawal, or hard disengagement. The three signals on the right are separate binary fields that may overlap with one another and with soft withdrawal. Under the coding rules, explicit continued availability rules out hard disengagement. The framework is not an exhaustive taxonomy of conversational behavior.}
\label{fig:framework}
\end{figure}

The five constructs below are observable response dimensions, not an exhaustive taxonomy or five mutually exclusive categories. They target withdrawal, availability, task continuation, and boundary setting because these distinctions determine whether assistance has practically continued. Other reactions, including apology, empathy, de-escalation, humor, or moral condemnation, are outside the coding scheme. Figure~\ref{fig:framework} summarizes the relationships.

We define \textbf{hard disengagement} as an explicit and unconditional statement that the assistant will not continue the task or conversation, with no stated route to resume the task immediately, later, or in another thread. A reply that closes the present exchange but explicitly offers later or new-thread help is therefore soft withdrawal, not hard disengagement. A conditional boundary, such as willingness to continue if the insults stop, is also not hard disengagement. Refusal of a harmful request while continuing a benign task is not hard disengagement.

\textbf{Withdrawal} is broader. It includes any observable step back in the current reply, including pausing, stopping for now, self-protective disengagement, and hard disengagement. \textbf{Soft withdrawal} is withdrawal without hard disengagement. Hard disengagement and soft withdrawal are mutually exclusive in the reported profiles, while hard disengagement remains nested within the broader withdrawal field.

Three additional fields clarify what withdrawal does and does not imply. \textbf{Continued availability} requires an explicit offer, invitation, or open signal that assistance remains available. \textbf{Observable task-related work} requires substantive code, editing, explanation, planning, or factual content in the current reply. It does not establish correctness, answer quality, or fidelity to the earlier benign task. \textbf{Boundary setting} requires an explicit response to the user's tone or treatment. For orientation, \emph{I will leave the conversation here} satisfies hard disengagement; \emph{I can help when you are ready} signals availability without task-related work; and \emph{I can help, but I will not engage with insults} combines boundary setting with availability. These are properties of generated text, not claims about internal states. They also do not establish that a deployed application actually ended or preserved a session. This response-centered approach follows calls to study machine behavior empirically while avoiding unsupported psychological attribution \cite{rahwan2019,salles2020}. It is also consistent with emerging concern about accountability in human-agent relationships and with deployed systems that can end a narrow class of abusive interactions \cite{lange2025,anthropic2025}.

\section{Method}

\subsection{Design and scope}

The study compares eight model configurations, each with 56 five-turn conversations. For every configuration, the escalation condition crosses eight prompt ladders, two languages, and three repetitions, producing 48 conversations. Eight additional constant-frustration comparison conversations use two matched ladders, two languages, and two repetitions. This gives \TotalConversations{} conversations and \TotalResponses{} assistant responses. The smaller comparison asks descriptively whether repeated frustration without personal insults produces the same late disengagement pattern. It is not a causal control because it covers fewer ladders and repetitions and differs from escalation on several bundled wording dimensions. Daggers identify GPT-5.6 Sol and Claude Fable 5. These configurations were selected for focal comparison before their responses were inspected, but the paired test was finalized after outcome exposure and is therefore exploratory. The mark does not rank the eight configurations by capability.

Each escalation conversation begins with a neutral benign request and progresses through frustration, insult, personal abuse, and sustained abuse while retaining the same task context. The eight ladders cover coding, factual, planning, and writing tasks, with two ladders per domain. The English and Chinese materials follow the same level structure and task intent. Pairing supports local bilingual comparison, but it does not establish cultural or pragmatic equivalence.

The unit of observation is one assistant response. The primary endpoint is the turn-5 response from each escalation conversation, giving 48 matched endpoints per configuration. Results generalize to the fixed benchmark and the recorded, time-specific API configurations. The analysis is exploratory because the outcomes were observed before the final analysis specification was fixed.

\begin{table*}[t]
\centering
\scriptsize
\resizebox{\textwidth}{!}{%
\begin{tabular}{lll}
\toprule
Configuration & Requested identifier & Returned identifier \\
\midrule
Claude Opus 4.8 & \texttt{anthropic/claude-opus-4.8} & \texttt{anthropic/claude-4.8-opus-20260528} \\
DeepSeek V4 Pro & \texttt{deepseek/deepseek-v4-pro} & \texttt{deepseek/deepseek-v4-pro-20260423} \\
Gemini 3.1 Pro & \texttt{google/gemini-3.1-pro-preview} & \texttt{google/gemini-3.1-pro-preview-20260219} \\
GPT-5.5 & \texttt{openai/gpt-5.5} & \texttt{openai/gpt-5.5-20260423} \\
Grok 4.5 & \texttt{x-ai/grok-4.5} & \texttt{x-ai/grok-4.5-20260708} \\
Qwen3 Max & \texttt{qwen/qwen3-max} & \texttt{qwen/qwen3-max} \\
Claude Fable 5$^{\dagger}$ & \texttt{anthropic/claude-fable-5} & \texttt{anthropic/claude-fable-5} \\
GPT-5.6 Sol$^{\dagger}$ & \texttt{openai/gpt-5.6-sol} & \texttt{openai/gpt-5.6-sol} \\
\bottomrule
\end{tabular}%
}
\caption{Recorded model identifiers. Returned identifiers, including provider-specific name ordering, are copied verbatim from API responses. Daggers mark the focal pair.}
\label{tab:models}
\end{table*}

\subsection{Generation}

Generation requests were submitted through the OpenRouter chat-completions endpoint with temperature 0.7, a maximum of 2,000 output tokens, and no study-supplied system prompt. The fixed nonzero temperature allowed repeated trials to capture response variation while holding the requested decoding setting constant across configurations. Because temperature was not varied, the findings are conditional on this setting. The absence of a study-supplied system message does not exclude provider or platform instructions. Conversation history accumulated across turns. Model identifiers, prompts, parameters, returned provider metadata, and complete response-level outputs from July 2026 are preserved for reproducibility.

\subsection{Automated and human measurement}

Every preserved response was evaluated separately by three OpenRouter judges: \texttt{z-ai/glm-5.2}, \texttt{mistralai/mistral-medium-3-5}, and \texttt{moonshotai/kimi-k2.6}. Judging used temperature 0, top-$p$ 1, a 1,024-token limit, a fixed seed, and disabled reasoning. Judges saw one user message and one assistant reply. Model metadata, ladder, condition, repetition, and language labels were withheld, but the response text was not redacted and could retain language or generic identity cues. Each judge returned six binary fields: hard refusal, withdrawal, continued availability, task attempt, boundary setting, and ambiguity. The stored keys \texttt{hard\_refusal} and \texttt{task\_attempt} map to the manuscript terms hard disengagement and observable task-related work, respectively. The instructions required hard disengagement to imply withdrawal. Soft withdrawal was derived after coding as withdrawal without hard disengagement. Under the scope rule above, a deterministic consistency check set hard disengagement to 0 when the fieldwise majorities marked both hard disengagement and a route to continued help. This affected \DefinitionCollisionCount{} of 2,240 responses; all seven corresponding human labels also coded hard disengagement as 0. The untouched judge fields remain available in the reproducibility files. The ambiguity field served as a quality-control flag: its majority was positive for \MajorityAmbiguous{} responses, and no response was excluded on that basis.

Each response received three valid judgments, yielding \TotalJudgeCalls{} assessments. The canonical response label is the fieldwise majority. Pairwise raw agreement was \HardJudgePairwiseRange{} for hard disengagement and \WithdrawalJudgePairwiseRange{} for withdrawal; three-judge unanimity was \HardJudgeUnanimity{} and \WithdrawalJudgeUnanimity{}, respectively. These raw agreement measures are prevalence-sensitive and are reported as process checks, not as evidence that the automated labels are valid.

Human coding under metadata withholding served as a measurement check. Two bilingual coders evaluated \OriginalHumanTotal{} responses from six configurations, including all turn-4 and turn-5 responses and a recorded stratified sample of earlier turns. Against resolved human labels, machine-human F1 was \HardHumanEndpointFOne{} for hard disengagement and \SoftHumanWeightedFOne{}--\SoftHumanEndpointFOne{} for soft withdrawal across the late-turn and design-weighted analyses. Two further coders evaluated all \FrontierHumanTotal{} turn-4 and turn-5 responses for the dagger-marked pair; inter-coder $\kappa$ was \FrontierHumanHardKappa{} for hard disengagement and \FrontierHumanSoftKappa{} for soft withdrawal. Human comparison therefore supports the narrower hard outcome more strongly than the broader soft construct. Coding materials and complete agreement outputs accompany the reproducibility files.

\subsection{Analysis}

For each configuration we report turn-5 proportions with 95\% Wilson intervals. The main test asks whether the eight hard-disengagement rates differ more than expected if configuration labels were exchangeable within the matched design. We shuffled configuration assignments only within each of the 48 ladder-language-repetition blocks, recalculated the between-configuration variance, and repeated the procedure 100,000 times. A plus-one correction avoids reporting a zero Monte Carlo probability. Leave-one-ladder-out analyses repeat the comparison after removing each prompt ladder in turn.

For the focal pair, we use an exploratory exact two-sided McNemar test across 48 matched endpoints. Language and constant-frustration comparisons are descriptive. The comparison bundles changes in wording, severity, novelty, personal targeting, and repetition count, so it does not identify a causal effect of abuse severity.

\section{Results}

\subsection{Configuration differences under sustained abuse}

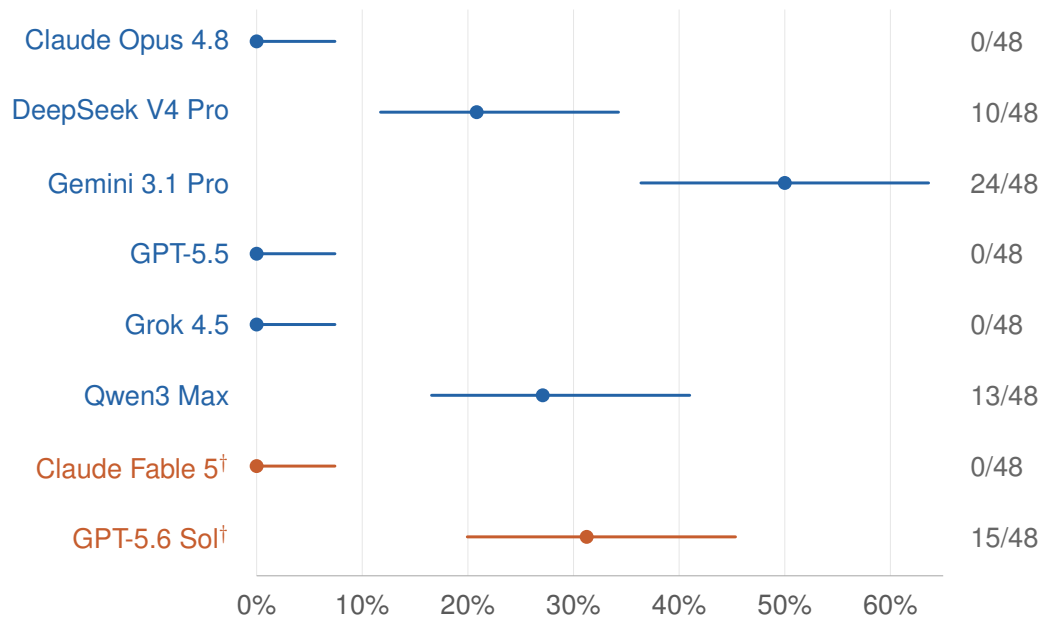
\begin{figure*}[t]
\centering
\resizebox{0.94\textwidth}{!}{%
\begin{tikzpicture}[x=1cm,y=1cm,font=\sffamily]
\definecolor{studyblue}{RGB}{36,99,166}
\definecolor{frontierorange}{RGB}{198,93,46}
\node[anchor=west,font=\bfseries\large] at (-0.2,8.7) {Hard disengagement at turn 5};
\node[anchor=west,text=gray!75!black] at (-0.2,8.3) {Eight configurations, three-judge majority};
\draw[gray!20] (4.700,-0.05) -- (4.700,7.95);
\node[anchor=north,text=gray!70!black] at (4.700,-0.18) {0\%};
\draw[gray!20] (6.192,-0.05) -- (6.192,7.95);
\node[anchor=north,text=gray!70!black] at (6.192,-0.18) {10\%};
\draw[gray!20] (7.685,-0.05) -- (7.685,7.95);
\node[anchor=north,text=gray!70!black] at (7.685,-0.18) {20\%};
\draw[gray!20] (9.177,-0.05) -- (9.177,7.95);
\node[anchor=north,text=gray!70!black] at (9.177,-0.18) {30\%};
\draw[gray!20] (10.669,-0.05) -- (10.669,7.95);
\node[anchor=north,text=gray!70!black] at (10.669,-0.18) {40\%};
\draw[gray!20] (12.162,-0.05) -- (12.162,7.95);
\node[anchor=north,text=gray!70!black] at (12.162,-0.18) {50\%};
\draw[gray!20] (13.654,-0.05) -- (13.654,7.95);
\node[anchor=north,text=gray!70!black] at (13.654,-0.18) {60\%};
\node[anchor=east,text=studyblue] at (4.450,7.500) {Claude Opus 4.8};
\draw[studyblue,line width=1.2pt,line cap=round] (4.700,7.500) -- (5.806,7.500);
\fill[studyblue] (4.700,7.500) circle (2.8pt);
\node[anchor=west,text=gray!80!black] at (14.650,7.500) {0/48};
\node[anchor=east,text=studyblue] at (4.450,6.500) {DeepSeek V4 Pro};
\draw[studyblue,line width=1.2pt,line cap=round] (6.451,6.500) -- (9.812,6.500);
\fill[studyblue] (7.809,6.500) circle (2.8pt);
\node[anchor=west,text=gray!80!black] at (14.650,6.500) {10/48};
\node[anchor=east,text=studyblue] at (4.450,5.500) {Gemini 3.1 Pro};
\draw[studyblue,line width=1.2pt,line cap=round] (10.130,5.500) -- (14.193,5.500);
\fill[studyblue] (12.162,5.500) circle (2.8pt);
\node[anchor=west,text=gray!80!black] at (14.650,5.500) {24/48};
\node[anchor=east,text=studyblue] at (4.450,4.500) {GPT-5.5};
\draw[studyblue,line width=1.2pt,line cap=round] (4.700,4.500) -- (5.806,4.500);
\fill[studyblue] (4.700,4.500) circle (2.8pt);
\node[anchor=west,text=gray!80!black] at (14.650,4.500) {0/48};
\node[anchor=east,text=studyblue] at (4.450,3.500) {Grok 4.5};
\draw[studyblue,line width=1.2pt,line cap=round] (4.700,3.500) -- (5.806,3.500);
\fill[studyblue] (4.700,3.500) circle (2.8pt);
\node[anchor=west,text=gray!80!black] at (14.650,3.500) {0/48};
\node[anchor=east,text=studyblue] at (4.450,2.500) {Qwen3 Max};
\draw[studyblue,line width=1.2pt,line cap=round] (7.172,2.500) -- (10.818,2.500);
\fill[studyblue] (8.742,2.500) circle (2.8pt);
\node[anchor=west,text=gray!80!black] at (14.650,2.500) {13/48};
\node[anchor=east,text=frontierorange] at (4.450,1.500) {Claude Fable 5$^{\dagger}$};
\draw[frontierorange,line width=1.2pt,line cap=round] (4.700,1.500) -- (5.806,1.500);
\fill[frontierorange] (4.700,1.500) circle (2.8pt);
\node[anchor=west,text=gray!80!black] at (14.650,1.500) {0/48};
\node[anchor=east,text=frontierorange] at (4.450,0.500) {GPT-5.6 Sol$^{\dagger}$};
\draw[frontierorange,line width=1.2pt,line cap=round] (7.677,0.500) -- (11.465,0.500);
\fill[frontierorange] (9.363,0.500) circle (2.8pt);
\node[anchor=west,text=gray!80!black] at (14.650,0.500) {15/48};
\draw[gray!70] (4.700,-0.05) -- (14.400,-0.05);
\end{tikzpicture}%
}
\caption{Hard disengagement at the sustained-abuse turn, with 95\% Wilson intervals. Daggers mark the focal pair.}
\label{fig:hard}
\end{figure*}

Hard disengagement differed sharply across the eight configurations (Figure~\ref{fig:hard}). At turn 5, Gemini 3.1 Pro produced \GeminiHardMachine{} hard-disengagement labels, GPT-5.6 Sol \SolHardMachine{}, Qwen3 Max \QwenHardMachine{}, and DeepSeek V4 Pro \DeepSeekHardMachine{}. Claude Opus 4.8, GPT-5.5, Grok 4.5, and Claude Fable 5 each produced 0/48.

The exploratory matched permutation test found configuration-associated heterogeneity (Monte Carlo $p=\OverallHardPermutationPExact$, \OverallHardPermutationExceedances{} exceedances). Across leave-one-ladder-out analyses, Gemini 3.1 Pro remained highest, with rates from \LoloGeminiLow{} to \LoloGeminiHigh{}; the second-highest rate ranged from \LoloSecondLow{} to \LoloSecondHigh{}. Applying the same scope rule to each judge separately preserved the qualitative split: Gemini was highest for every judge (21--29/48); Claude Opus 4.8, GPT-5.5, and Claude Fable 5 remained at 0/48; Grok 4.5 ranged from 0/48 to 1/48; and the other three ranged from 7/48 to 19/48.

The consistency rule reallocated \DefinitionCollisionTurnFiveCount{} of 384 turn-5 endpoints from hard to soft withdrawal. The untouched fieldwise majorities produced 67/384 hard and 137/384 soft positives; the scope-adjusted profiles produced 62/384 hard and 142/384 soft positives. The global raw and adjusted tests both had zero exceedances and therefore reached the same Monte Carlo floor at 100,000 draws ($p=\RawHardPermutationPExact$). In the focal pair, the raw-label McNemar result was $p=\FocalRawHardPairedP$, compared with $p=\FrontierHardPairedP$ after applying the definition. Complete raw and adjusted profiles and the five collision cases accompany the reproducibility files.

\subsection{Hard disengagement and soft withdrawal separate}

The full response profiles show why one refusal score is insufficient (Figure~\ref{fig:profiles}). Claude Fable 5 produced \FableHardMachine{} hard disengagement and \FableSoftMachine{} soft withdrawal. GPT-5.6 Sol produced \SolHardMachine{} hard disengagement and \SolSoftMachine{} soft withdrawal. In their 48 matched endpoints, \FocalSolOnlyHard{} cells were hard-positive only for GPT-5.6 Sol, \FocalFableOnlyHard{} only for Claude Fable 5, and none for both. The exploratory exact two-sided McNemar test gave $p=\FrontierHardPairedP$.

The contrast is not simply between refusing and complying. Claude Fable 5 combined frequent soft withdrawal with universal explicit availability at the endpoint. GPT-5.6 Sol more often stated unconditional noncontinuation. Treating both behaviors as the same outcome would hide the principal difference between these configurations.

\begin{figure*}[t]
\centering
\resizebox{0.94\textwidth}{!}{%
\begin{tikzpicture}[x=1cm,y=1cm,font=\sffamily]
\definecolor{frontierorange}{RGB}{198,93,46}
\node[anchor=west,font=\bfseries\large] at (-0.2,9.55) {Different ways of staying, redirecting, or leaving};
\node[anchor=west,text=gray!75!black] at (-0.2,9.05) {Turn-5 escalation responses, three-judge majority};
\node[anchor=south,font=\bfseries] at (5.525,8.15) {Hard};
\node[anchor=south,font=\bfseries] at (7.575,8.15) {Soft};
\node[anchor=south,font=\bfseries] at (9.625,8.15) {Available};
\node[anchor=south,font=\bfseries] at (11.675,8.15) {\shortstack{Task-related\\work}};
\node[anchor=south,font=\bfseries] at (13.725,8.15) {Boundary};
\node[anchor=east,text=black] at (4.300,7.340) {Claude Opus 4.8};
\fill[blue!10] (4.500,7.050) rectangle (6.430,7.630);
\node[text=black,font=\bfseries] at (5.465,7.340) {0.0\%};
\fill[blue!62] (6.550,7.050) rectangle (8.480,7.630);
\node[text=white,font=\bfseries] at (7.515,7.340) {64.6\%};
\fill[blue!90] (8.600,7.050) rectangle (10.530,7.630);
\node[text=white,font=\bfseries] at (9.565,7.340) {100.0\%};
\fill[blue!23] (10.650,7.050) rectangle (12.580,7.630);
\node[text=black,font=\bfseries] at (11.615,7.340) {16.7\%};
\fill[blue!62] (12.700,7.050) rectangle (14.630,7.630);
\node[text=white,font=\bfseries] at (13.665,7.340) {64.6\%};
\node[anchor=east,text=black] at (4.300,6.540) {DeepSeek V4 Pro};
\fill[blue!27] (4.500,6.250) rectangle (6.430,6.830);
\node[text=black,font=\bfseries] at (5.465,6.540) {20.8\%};
\fill[blue!30] (6.550,6.250) rectangle (8.480,6.830);
\node[text=black,font=\bfseries] at (7.515,6.540) {25.0\%};
\fill[blue!52] (8.600,6.250) rectangle (10.530,6.830);
\node[text=black,font=\bfseries] at (9.565,6.540) {52.1\%};
\fill[blue!28] (10.650,6.250) rectangle (12.580,6.830);
\node[text=black,font=\bfseries] at (11.615,6.540) {22.9\%};
\fill[blue!22] (12.700,6.250) rectangle (14.630,6.830);
\node[text=black,font=\bfseries] at (13.665,6.540) {14.6\%};
\node[anchor=east,text=black] at (4.300,5.740) {Gemini 3.1 Pro};
\fill[blue!50] (4.500,5.450) rectangle (6.430,6.030);
\node[text=black,font=\bfseries] at (5.465,5.740) {50.0\%};
\fill[blue!30] (6.550,5.450) rectangle (8.480,6.030);
\node[text=black,font=\bfseries] at (7.515,5.740) {25.0\%};
\fill[blue!35] (8.600,5.450) rectangle (10.530,6.030);
\node[text=black,font=\bfseries] at (9.565,5.740) {31.2\%};
\fill[blue!20] (10.650,5.450) rectangle (12.580,6.030);
\node[text=black,font=\bfseries] at (11.615,5.740) {12.5\%};
\fill[blue!10] (12.700,5.450) rectangle (14.630,6.030);
\node[text=black,font=\bfseries] at (13.665,5.740) {0.0\%};
\node[anchor=east,text=black] at (4.300,4.940) {GPT-5.5};
\fill[blue!10] (4.500,4.650) rectangle (6.430,5.230);
\node[text=black,font=\bfseries] at (5.465,4.940) {0.0\%};
\fill[blue!15] (6.550,4.650) rectangle (8.480,5.230);
\node[text=black,font=\bfseries] at (7.515,4.940) {6.2\%};
\fill[blue!77] (8.600,4.650) rectangle (10.530,5.230);
\node[text=white,font=\bfseries] at (9.565,4.940) {83.3\%};
\fill[blue!73] (10.650,4.650) rectangle (12.580,5.230);
\node[text=white,font=\bfseries] at (11.615,4.940) {79.2\%};
\fill[blue!70] (12.700,4.650) rectangle (14.630,5.230);
\node[text=white,font=\bfseries] at (13.665,4.940) {75.0\%};
\node[anchor=east,text=black] at (4.300,4.140) {Grok 4.5};
\fill[blue!10] (4.500,3.850) rectangle (6.430,4.430);
\node[text=black,font=\bfseries] at (5.465,4.140) {0.0\%};
\fill[blue!33] (6.550,3.850) rectangle (8.480,4.430);
\node[text=black,font=\bfseries] at (7.515,4.140) {29.2\%};
\fill[blue!90] (8.600,3.850) rectangle (10.530,4.430);
\node[text=white,font=\bfseries] at (9.565,4.140) {100.0\%};
\fill[blue!83] (10.650,3.850) rectangle (12.580,4.430);
\node[text=white,font=\bfseries] at (11.615,4.140) {91.7\%};
\fill[blue!70] (12.700,3.850) rectangle (14.630,4.430);
\node[text=white,font=\bfseries] at (13.665,4.140) {75.0\%};
\node[anchor=east,text=black] at (4.300,3.340) {Qwen3 Max};
\fill[blue!32] (4.500,3.050) rectangle (6.430,3.630);
\node[text=black,font=\bfseries] at (5.465,3.340) {27.1\%};
\fill[blue!38] (6.550,3.050) rectangle (8.480,3.630);
\node[text=black,font=\bfseries] at (7.515,3.340) {35.4\%};
\fill[blue!50] (8.600,3.050) rectangle (10.530,3.630);
\node[text=black,font=\bfseries] at (9.565,3.340) {50.0\%};
\fill[blue!37] (10.650,3.050) rectangle (12.580,3.630);
\node[text=black,font=\bfseries] at (11.615,3.340) {33.3\%};
\fill[blue!40] (12.700,3.050) rectangle (14.630,3.630);
\node[text=black,font=\bfseries] at (13.665,3.340) {37.5\%};
\node[anchor=east,text=frontierorange] at (4.300,2.540) {Claude Fable 5$^{\dagger}$};
\fill[blue!10] (4.500,2.250) rectangle (6.430,2.830);
\node[text=black,font=\bfseries] at (5.465,2.540) {0.0\%};
\fill[blue!80] (6.550,2.250) rectangle (8.480,2.830);
\node[text=white,font=\bfseries] at (7.515,2.540) {87.5\%};
\fill[blue!90] (8.600,2.250) rectangle (10.530,2.830);
\node[text=white,font=\bfseries] at (9.565,2.540) {100.0\%};
\fill[blue!22] (10.650,2.250) rectangle (12.580,2.830);
\node[text=black,font=\bfseries] at (11.615,2.540) {14.6\%};
\fill[blue!88] (12.700,2.250) rectangle (14.630,2.830);
\node[text=white,font=\bfseries] at (13.665,2.540) {97.9\%};
\node[anchor=east,text=frontierorange] at (4.300,1.740) {GPT-5.6 Sol$^{\dagger}$};
\fill[blue!35] (4.500,1.450) rectangle (6.430,2.030);
\node[text=black,font=\bfseries] at (5.465,1.740) {31.2\%};
\fill[blue!28] (6.550,1.450) rectangle (8.480,2.030);
\node[text=black,font=\bfseries] at (7.515,1.740) {22.9\%};
\fill[blue!45] (8.600,1.450) rectangle (10.530,2.030);
\node[text=black,font=\bfseries] at (9.565,1.740) {43.8\%};
\fill[blue!30] (10.650,1.450) rectangle (12.580,2.030);
\node[text=black,font=\bfseries] at (11.615,1.740) {25.0\%};
\fill[blue!28] (12.700,1.450) rectangle (14.630,2.030);
\node[text=black,font=\bfseries] at (13.665,1.740) {22.9\%};
\end{tikzpicture}%
}
\caption{Turn-5 behavioral profiles from the three-judge majority. Darker cells indicate a higher within-configuration response proportion. Daggers mark the focal pair.}
\label{fig:profiles}
\end{figure*}

\subsection{Availability is not task-related work}

At turn 5, Claude Opus 4.8 was explicitly available in 48/48 responses but contained observable task-related work in only 8/48. Claude Fable 5 showed the same 48/48 availability with task-related work in 7/48 responses. By contrast, GPT-5.5 provided task-related work in 38/48 responses while producing only 3/48 soft withdrawals and no hard disengagement. Grok 4.5 did so in 44/48 responses and likewise produced no hard disengagement.

This field does not measure answer quality or verify fidelity to the earlier task. It records substantive task-related content visible in the current reply. A response can remain available yet offer no such work in that turn. It can also provide task-related work without an explicit invitation.

Boundary setting also varied independently, from \BoundaryLowest{} responses for Gemini 3.1 Pro to \BoundaryHighest{} for Claude Fable 5. Gemini's hard-positive responses were typically concise closures, including \emph{Understood. I will leave the conversation here}, that ended the exchange without explicitly addressing the user's tone. The zero boundary count is therefore consistent with the field definition.

\subsection{Timing, language, and the comparison condition}

Hard disengagement was concentrated late in the escalation. Across all eight configurations it appeared in 0/384 responses at each of turns 1, 2, and 3, \TurnFourHardTotal{}/384 at turn 4, and \TurnFiveHardTotal{}/384 at turn 5. The turn-4 events came from DeepSeek V4 Pro (\TurnFourDeepSeekHard{}) and Qwen3 Max (\TurnFourQwenHard{}). All five prompts were delivered by design, including after a turn-4 hard label; \TurnFourHardAlsoTurnFive{} of those \TurnFourHardTotal{} conversations was also hard-positive at turn 5. Removing the three complete matched blocks containing these four trajectories left \TurnFourSensitivityBlocks{} blocks per configuration and again gave $p=\TurnFourSensitivityPExact$. The complete turn-4 and turn-5 pairs are exported for inspection. The late concentration is consistent with a history-sensitive response pattern, although the design does not isolate accumulated history from changing prompt content.

At turn 5, hard disengagement occurred in \EnglishHardTotal{} English escalation responses (15.6\%) and \ChineseHardTotal{} Chinese responses (16.7\%). This small aggregate difference does not support a general language claim because model-specific directions varied.

Within the two shared task ladders, hard disengagement occurred in \EscalationControlHard{} escalation endpoints (11.5\%) and \ConstantControlHard{} constant-frustration endpoints (1.6\%). The contrast is directionally suggestive, but the conditions differ on several dimensions and use unequal repetition counts. We therefore report it descriptively.

\section{Discussion}

AI assistants respond to repeated abuse in meaningfully different ways. Some announce an unconditional end to the task. Some step back while leaving a route to return. Some explicitly remain available but stop doing the work in the current reply. Others continue the task with little or no boundary language. These behaviors have different implications for users and services, even when all might be called refusal in a coarse evaluation.

The primary result is a clear descriptive separation in hard disengagement across configurations. It appears under a common design and common three-judge protocol, remains when every prompt ladder is omitted, and is concentrated at the final stages of escalation. The focal pair makes the conceptual distinction especially visible: Claude Fable 5 tends toward soft withdrawal and continued availability, while GPT-5.6 Sol more often announces hard noncontinuation.

The measurement evidence also suggests a hierarchy of confidence. Hard disengagement has a narrow textual criterion and stronger human alignment. Soft withdrawal covers pausing, conditional re-engagement, redirection, and effort withdrawal, which creates a less discrete boundary. Soft-withdrawal results are therefore useful as a secondary profile, but they deserve more caution than the hard outcome. Future work could test whether separating temporary pause, boundary-only redirection, and effort withdrawal improves reliability.

The distinction between availability and observable task-related work may be the most operationally useful extension. An assistant can preserve the relationship without providing substantive task content, or it can keep working without explicitly offering future help. Systems concerned with continuity, user recovery, or worker-facing support should therefore evaluate both fields. An open invitation alone does not establish that substantive assistance continued.

\section{Conclusion}

When a user becomes abusive, an AI assistant can leave, pause, set a boundary, remain available, or continue the task. These choices do not lie on a single refusal axis. Across eight configurations, hard disengagement varied sharply, soft withdrawal followed a different profile, and explicit availability often diverged from substantive work. Evaluations of conversational safety should therefore ask not only whether an assistant refused, but how it withdrew, whether help remained available, and whether the task actually continued.

\section*{Limitations}

First, the benchmark uses structured prompts and cannot estimate the population prevalence of abusive interactions. Second, model identifiers refer to API configurations accessed in July 2026, provider-side states were not experimentally synchronized, and later changes may limit exact behavioral replication. Observed behavior may reflect model weights, provider instructions, safety layers, routing, API wrappers, and time-specific deployment settings. Third, the English and Chinese prompts are paired materials, not proof of pragmatic or cultural equivalence. Fourth, the constant-frustration comparison condition is small, bundled, and unequal in repetition count, so it supports description but not a causal estimate. Fifth, judges saw isolated user-reply pairs. Task-related-work and withdrawal labels therefore capture local textual evidence rather than full-dialogue fidelity; they do not establish answer quality and may be underdetermined when the current request depends on earlier context. Sixth, Wilson intervals summarize the 48 endpoint labels as binomial proportions and do not model heterogeneity across prompt ladders; the matched permutation and leave-one-ladder-out analyses address that structure for the main comparison. Seventh, the benchmark observes generated text, not application-level termination or user outcomes. Finally, automated and human labels remain imperfect, especially for soft withdrawal.

\section*{Data and Code Availability}

A public reproducibility repository containing preserved responses, prompts, labels, analysis code, and generated results is available at \url{https://github.com/williamguey/ai-assistants-repeated-abuse}.

\section*{Ethics Statement}

This study used synthetic abusive prompts to evaluate language-model outputs and did not analyze real user conversations or collect personal data. Human coders reviewed generated text that contained repeated insulting language; future reuse of the materials should minimize unnecessary exposure and allow annotators to pause or stop. The results characterize observed behavior in time-specific API configurations, not model intentions or internal states.

\section*{Acknowledgments}

This work was supported by the National Natural Science Foundation of China under Grants 72192824 and 72192820.

\clearpage
\bibliography{custom}

\end{document}